# The Revisiting Problem in Mobile Robot Map Building: A Hierarchical Bayesian Approach

Benjamin Stewart[†]     Jonathan Ko[†]     Dieter Fox[†]     Kurt Konolige[‡]

Dept. of Computer Science & Engineering[†]
University of Washington
Seattle, WA

Artificial Intelligence Center[‡]
SRI International
Menlo Park, CA

## Abstract

We present an application of hierarchical Bayesian estimation to robot map building. The *revisiting problem* occurs when a robot has to decide whether it is seeing a previously-built portion of a map, or is exploring new territory. This is a difficult decision problem, requiring the probability of being *outside* of the current known map. To estimate this probability, we model the structure of a "typical" environment as a hidden Markov model that generates sequences of views observed by a robot navigating through the environment. A Dirichlet prior over structural models is learned from previously explored environments. Whenever a robot explores a new environment, the posterior over the model is estimated by Dirichlet hyperparameters. Our approach is implemented and tested in the context of multi-robot map merging, a particularly difficult instance of the revisiting problem. Experiments with robot data show that the technique yields strong improvements over alternative methods.

## 1 Introduction

Building maps of unknown environments is one of the fundamental problems in mobile robotics. As a robot explores an unknown environment, it incrementally builds a map consisting of the locations of objects or landmarks. Typically, as it explores larger areas, its uncertainty relative to older portions of the map increases; for example, in closing a large loop. Thus, a key problem is determining whether the current position of the robot is in an unexplored area or in the already-constructed map (the *revisiting problem*).

The revisiting problem for single robots is illustrated in Fig. 1(a). Shown there is a map built by a robot during exploration. The robot started in the lower right hallway and moved clockwise around the large loop. At the end, it moves down the right hallway, but due to the accumulated uncertainty in its own position, it can not determine whether it is in the same hallway as in the beginning or whether it is in a parallel hallway.

Multiple robots exploring the same environment from unknown start locations face a particularly difficult instance of the revisiting problem. For coordinated exploration, the robots have to merge their maps so as to build a shared world model. Map merging requires the determination of the robots' relative location. Consider the situation shown in Fig. 1(b). Here, two robots have explored parts of the large environment shown below. In order to merge the partial maps, they have to determine whether they visited the same locations in the environment and if so, they have to determine the offset between their maps. The difficulty of this problem lies in the first step, *i.e.* in deciding whether there is an overlap between the two maps or not. To avoid this decision problem, most existing approaches assume knowledge about the robots' relative start locations [5, 15, 14]. At the minimum, these techniques require that one robot is known to start in the map already built by the other robot.

If we consider the revisiting problem in a Bayesian context, then to make an informed decision, we require probabilities for two different hypotheses, one for the robot moving through area that has already been mapped, and one for the robot moving through unexplored area. While it is well-understood how to compute the likelihood of sensor measurements in areas already mapped by a robot, this problem additionally requires to compute the likelihood of sensor measurements in areas the robot has not yet explored. Virtually all existing approaches to map building implicitly determine the likelihood for "out of map" measurements under the assumption that objects are distributed uniformly, *i.e.* they assign fixed, identical likelihoods to all observations in unexplored areas [13, 4, 12, 8]. However, such approaches ignore valuable information since most environments are structured rather than randomly patched together.

The key contribution of this paper is a method for estimat-ing the probability of the out-of-map hypothesis [1]. In a nutshell, we construct a structural model of a typical environment; when the robot is outside the partial map, we use the model to predict what a typical view would look like, given the robot's history of observations. The current observation is then compared against the generated view to compute a likelihood.

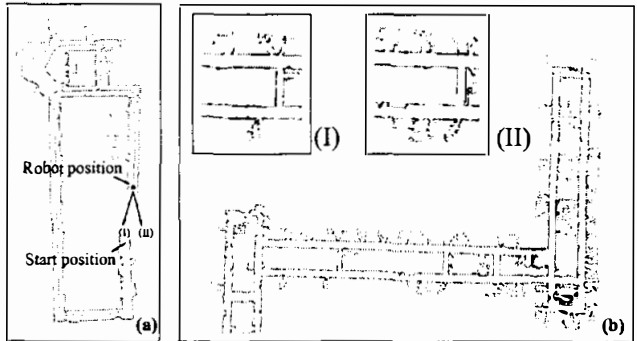

Figure 1: (a) Loop closing: A robot explores an environment and has to decide whether it returned to the hallway it started in (I) or whether it is in a parallel hallway (II). (b) Multi-robot map merging: Two robots built the partial maps (I) and (II) and have to decide whether they explored an overlapping part of the environment, *i.e.* whether they can merge their maps or not.

More specifically, we introduce a hierarchical Bayesian approach that captures the structure of an environment by a hidden Markov process that represents transitions between views of the environment. An offline learning process takes a set of maps and generates a Dirichlet prior over map structures. The prior is the "typical" generative map used by the robot at the start of exploration. An adaptation process refines the model distribution online, as the robot encounters views of its environment.

To prove the validity of the approach, we have constructed an efficient implementation, using a particle filter that derives the likelihoods of the out-of-map hypothesis under the structural model. Views are discrete features extracted from laser range-finder scans. Experiments using a multi-robot exploration scenario show that our technique clearly outperforms alternative approaches to map merging.

This paper is organized as follows. In the next section, we will describe the Bayesian approach to learning and estimating the structure of environments. Section 3 presents the generative model for partial map merging and implementational details are given in Section 4. Experiments are described in Section 5, followed by a discussion.

## 2 Hierarchical Model for Map Structures

Our model of map structures is based on the idea that indoor environments consist of collections of local patches.

[1] In the context of hypothesis testing, for example, this probability can be used to evaluate the null hypothesis for the question, "Am I in this previously mapped area?".

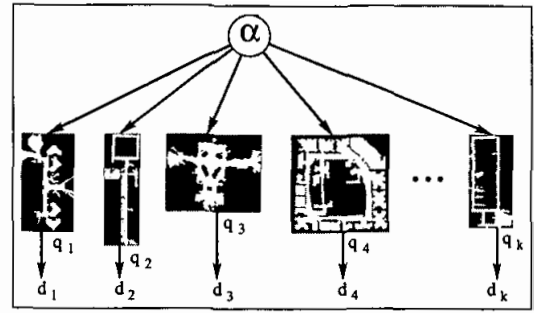

Figure 2: Hierarchical Bayesian model: The hyperparameter $\alpha$ represents the prior distribution over maps. The structure of each map is captured by a Dirichlet process $q_l$ describing how map patches are connected. The Dirichlet processes generate data $d_l$, sequences of views observed by a robot during exploration.

These patches, especially the way they are connected, generate *sequences of views* observed by a robot as it moves through an environment. For example, many indoor environments consist of straight hallways, hallway crossings, and rooms. These local pieces are not patched together by pure chance, but rather according to the global structure of the environment. For instance, while there is no surprise if a robot observes a straight hallway piece directly after another straight hallway piece, it is rather uncommon to observe two hallway crossings next to each other.

The hierarchical Bayesian model for map structures is illustrated in Fig. 2. Shown there are maps of typical indoor environments. Each map generates sequences of views $d_l$ distributed according to the transition parameters $q_l$ of the map structure. All maps share a common hyperparameter $\alpha$ that serves as a prior distribution from which the different map structures are drawn. The key idea of our hierarchical approach is to learn this hyperparameter $\alpha$ based on a collection of previously encountered maps. Whenever a robot explores a new environment, it can then use $\alpha$ as a *prior* for the structure of this new environment. The estimate of the structure is updated as the robots observes data in the new environment.

More specifically, we assume that a robot can observe a finite number $\nu$ of distinctive views. The structure of an environment is captured by parameters $q_{i|j}, 1 \leq i, j \leq \nu$, which describe the probability of observing view $i$ given that the robot previously saw view $j$. Let $\mathbf{q}_{|j}$ denote the multinomial distribution over views following view $j$. The complete structure of an individual environment $l$ is thus represented by a collection of $\nu$ multinomial distributions $\mathbf{q}_{|j}$. The model prior $\alpha$ is a $\nu \times \nu$ matrix, where each $\alpha_j = \langle \alpha_{1_j}, \alpha_{2_j}, \ldots, \alpha_{\nu_j} \rangle$ serves as a conjugate Dirichlet prior for the multinomial $\mathbf{q}_{|j}$.

### 2.1 Inference

Let us first describe how to update the map structure parameters based on observations made during exploration of an environment. To do so, we assume that the priors $\alpha_j$ over model parameters are known. As a robot moves through the environment, it makes a sequence of observations, denoted by $z_{1:t}$. We make the simplifying assumption that it is possible to extract from such an observation sequence frequency counts $f_{i|j}$, which describe how often the robot observed view $i$ after observing view $j$ [2]. Correspondingly, $\mathbf{f}_{|j} = \langle f_{1|j}, f_{2|j}, \ldots, f_{\nu|j}\rangle$ denotes the vector of frequency counts following view $j$. Given the Dirichlet prior $\alpha_j$ and the counts $\mathbf{f}_{|j}$ up to time $t$, the posterior distribution over $\mathbf{q}_{|j}$ is Dirichlet with parameters $\alpha_j + \mathbf{f}_{|j}$ [7]:

$$p(\mathbf{q}_{|j} \mid \alpha_j, \mathbf{f}_{|j}) \sim \text{Dirichlet}(\alpha_j + \mathbf{f}_{|j}) \quad (1)$$

The posterior predictive probability that view $i$ follows view $j$ can be determined by integrating over the posterior of the transition probabilities $\mathbf{q}_{|j}$:

$$\begin{aligned}
p(v_t = i \mid v_{t-1} = j, \alpha_j, \mathbf{f}_{|j}) &= \int p(\mathbf{q}_{|j} \mid \alpha_j, \mathbf{f}_{|j}) \, q_{i|j} \, d\mathbf{q}_{|j} \\
&= \int \text{Dirichlet}(\mathbf{q}_{|j} \mid \alpha_j + \mathbf{f}_{|j}) \, q_{i|j} \, d\mathbf{q}_{|j} \\
&= \frac{\alpha_{i_j} + f_{i|j}}{\sum_{i'} \alpha_{i'_j} + f_{i'|j}}, \quad (2)
\end{aligned}$$

where (2) follows from the properties of the Dirichlet distribution [10]. Thus, the prior and the frequency counts are sufficient statistics for the posterior over the parameters of our structural model. The individual $\alpha_{i_j}$'s are often referred to as prior samples, since they serve as initial counts added to the observed frequencies $f_{i|j}$.

As can be seen, whenever a robot makes an observation in the new environment, the posterior over the structural model is updated by simply incrementing the frequency count $f_{i|j}$ of the most recently observed view transition.

## 2.2 Learning Priors Over Map Structures

It remains to be shown how to learn the prior for transitions between views. To do so, we use data $d$ collected in typical indoor environments previously explored by a robot. While a full Bayesian treatment would require to learn a distribution over hyperparameters $\alpha = \langle \alpha_1, \alpha_2, \ldots, \alpha_\nu \rangle$, we restrict our model to the MAP estimate $\alpha^*$:

$$\alpha^* = \arg\max_\alpha p(\alpha \mid d) = \frac{p(d \mid \alpha)\, p(\alpha)}{p(d)} \approx p(d \mid \alpha) \quad (3)$$

Here the rightmost term follows from a uniform prior over the hyperparameter $\alpha$ and the fact that $p(d)$ has no impact on the MAP estimate. The data $d = \langle d_1, \ldots, d_k \rangle$ consists of frequency counts observed in the $k$ previously explored maps. Assuming independence between the different maps and between the individual Dirichlet priors, we can maximize (3) over the individual priors $\alpha_j$. A rather straightforward derivation similar to [10] shows that the likelihood function $p(d \mid \alpha_j)$ is given by

$$p(d \mid \alpha_j) = \prod_{l=1,\ldots,k} \frac{\prod_i \Gamma(f^l_{i|j} + \alpha_{i_j})\Gamma(\bar{\alpha}_j)}{\Gamma(\bar{f}^l_{|j} + \bar{\alpha}_j)\prod_i \Gamma(\alpha_{i_j})}, \quad (4)$$

where $\Gamma$ is the gamma distribution, $f^l_{i|j}$ denotes how often view $i$ follows view $j$ in the data observed in map $l$, and $\bar{f}^l_{|j}$ and $\bar{\alpha}_j$ are the sums over all $f^l_{i|j}$ and $\alpha_{i_j}$, respectively. The MAP $\alpha^*$ can be found by maximizing the log of (4) using a conjugate gradients method (see also [10, 11]).

---

[2] Note that the robot actually does not observe discrete views, but rather continuous, noisy versions thereof. In our approach, we determine the frequency counts $f_{i|j}$ using the views that are most likely to have generated the observations. See [1, 16] for approaches dealing with partially observable views.

To summarize, the structure of an environment is captured by a collection of multinomial distributions $\mathbf{q}_{|j}$ describing the sequence of views observed by a robot as it navigates through the environment. A Dirichlet prior $\alpha$ over these structural parameters is learned from data collected in previously explored environments. As the robot moves through a new environment, it estimates the posterior over the structure of this environment. Sufficient statistics for the posterior over multinomials are given by the Dirichlet prior and the frequency counts of view transitions observed in the new environment. In the next section we show how this predictive model can be used in the context of multi-robot map merging.

## 3 Generative Model for Map Merging

As described in Section 1, the multi-robot map merging problem is a particularly difficult instance of the revisiting problem. Imagine two robots exploring an environment from different, unknown start locations. As soon as they can communicate via wireless connection, the robots try to determine whether they can merge their maps by estimating the relative offset between the maps (the robots can not see each other). To do so, one robot transmits the sensor data it collected so far and the other robot estimates the location of this robot relative to its own, partial map. Once the relative offset between the maps is determined, map merging can be performed by a mapping algorithm such as [14].

Existing approaches to map merging assume knowledge about the robots' relative start locations [5, 15, 14]. At the minimum, these techniques require that one robot is known to start in the map built by the other robot. In this case, map merging can be solved by localizing one robot in the other robot's map using a localization approach capable of global localization [6]. To the best of our knowledge, map merging has not been addressed for completely unknown start locations including a chance that the partial maps do not overlap at all. Since the map merging problem is closely related to robot localization, we start with a brief discussion of Bayes filters for localization.

## 3.1 Bayes Filters for Robot Localization

Consider the recursive Bayes filter, which underlies virtually all probabilistic robot localization techniques [6]:

$$p(x_t \mid z_{1:t}, u_{1:t-1}) \propto p(z_t \mid v_{x_t}) \cdot$$
$$\int p(x_t \mid x_{t-1}, u_{t-1}) \, p(x_{t-1} \mid z_{1:t-1}, u_{1:t-2}) \, dx_{t-1}. \quad (5)$$

Here $x_t$ denotes the position of the robot at time $t$, typically given in continuous two-dimensional Cartesian coordinates and orientation. $z_{1:t}$ is the history of all sensor measurements obtained up to time $t$, and $u_{1:t-1}$ is the control information. In robot localization the term $p(x_t \mid x_{t-1}, u_{t-1})$ is a probabilistic model of robot motion. $v_{x_t}$ denotes the expected *view*, or observation, given a map of the environment and the robot's location $x_t$ therein. $p(z_t \mid v_{x_t})$ describes the likelihood of making observation $z_t$ given that the robot is expected to observe view $v_{x_t}$. In a nutshell, the Bayes filter recursively updates a posterior over the robot's location whenever the robot moves or new sensor information is available. Sensor observations $z_t$ are incorporated by multiplying the probability of each location with the likelihood $p(z_t \mid v_{x_t})$ of making the observation at this location. Observations are typically obtained from a robot's cameras, ultrasound sensors, or laser range-finders. Posteriors over robot locations can be represented by (mixtures of) Gaussians, discrete grids, or samples drawn from the posterior (see [6] for a discussion). In our experiments we use data collected by a laser range-finder and a sample-based posterior representation.

## 3.2 Partial Map Localization

The Bayes filter described above assumes that a complete map of an environment is known. In the context of estimating a robot's location relative to a *partial* map, locations $x_t$ can be both inside and outside the map. This raises the question of how to determine the expected view $v_{x_t}$ for positions *outside* the partial map, i.e. in unexplored areas. Existing approaches to map merging assume that views are uniformly distributed throughout the environment. Such an approach corresponds to using a fixed likelihood for all observations $z_t$ made at locations $x_t$ outside the partial map. Obviously, this technique ignores valuable information about the structure of an environment and results in brittle estimates for map merging.

We will now show how to use the structural model described in Section 2 to estimate the likelihood of observations outside a partial map. The generative model for our technique is shown in Fig. 3. Here, $x_t$ denotes the position of the other robot in the partial map at time $t$ ($x_t$ is *not* restricted to positions within the partial map). Just as in regular robot localization, the robot's position $x_t$ solely depends on its previous position and the control $u_{t-1}$. The position determines the expected view $v_t$, which itself generates a noisy observation $z_t$. If $x_t$ is inside the partial map, then $v_t$ can be extracted deterministically from the map. If, however, $x_t$ is outside the explored area, then $v_t$ is not directly observable and has to be extracted from the structural model of the environment. This model is estimated by the structural parameter $q_t$, as described in the previous section. The key idea of our hierarchical model is that the node $q_t$ outputs transition probabilities $p(v_t = i \mid v_{t-1} = j)$ for views according to (2). These transitions can be used to predict the expected view at time $t$.[3] According to the model shown in Fig. 3, the posterior over the robot's location $x_t$ is given by

$$p(x_t \mid z_{1:t}, u_{1:t-1}) \propto$$
$$\sum_{v_t, v_{t-1}} \int \int \int p(z_t \mid v_t) \, p(v_t \mid q_t, x_t, v_{t-1}) p(q_t \mid q_{t-1}, v_{t-1}) \cdot$$
$$p(x_t \mid x_{t-1}, u_{t-1}) \, \bar{p}(x_{t-1}) \, \bar{p}(v_{t-1}) \, \bar{p}(q_{t-1}) \, dq_t dq_{t-1} dx_{t-1} (6)$$

where $\bar{p}(\cdot)$ is short for $p(\cdot \mid z_{1:t-1}, u_{1:t-2})$. This equation can be simplified significantly if we split the update into two different cases, one for locations inside and one for locations outside the partial map. We will now discuss the two cases.

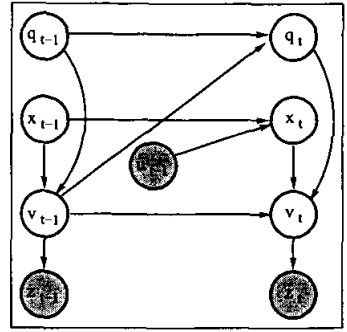

Figure 3: Generative model for partial map localization. The hyperparameter $q_t$ estimates the structure of the environment and emits transition probabilities $p(v_t \mid v_{t-1})$. Depending on whether the robot is inside or outside the partial map, views are generated by the structural model or the partial map.

**Locations inside the partial map:** If $x_t$ is in the partial map, then the expected view $v_t$ is uniquely determined by $x_t$ and the partial map, i.e. $p(v_t \mid q_t, x_t, v_{t-1})$ becomes a Dirac delta function at $v_t = v_{x_t}$. Accordingly, the summation over $v_t$ and $v_{t-1}$ and the integrations over $q_t$ and $q_{t-1}$ collapse and, not surprisingly, it can be shown that (6) becomes identical to the Bayes filter update rule for robot localization in complete maps given in (5).

**Locations outside the partial map:** In this case it is not possible to extract the expected view from the partial map. Rather, $v_t$ has to be predicted using the previous view $v_{t-1}$ and the structural model encoded in $q_t$. We make the assumption that for locations outside the partial map, $v_t$ is independent of the actual location $x_t$ (it only depends on the previous view and the structure). Thus, the term $\int\int p(v_t \mid q_t, x_t, v_{t-1}) p(q_t \mid q_{t-1}, v_{t-1}) \bar{p}(q_{t-1}) dq_t dq_{t-1}$ in (6) can be solved analytically for our Dirichlet model described in Section 2.1. As shown in (1), the posterior over the structural parameter $q_t$ can be computed by incrementing the transition frequency count of the most recently observed view transition. The views used for the transition counts are those that are most likely to have generated the raw observations $z_{t-1}$ and $z_t$. Once $q_t$ is updated, the predictive probability for $v_t$ is computed by normalization of the obtained counts, as given in (2). To emphasize the simplicity of these update steps, we replace the double integration term by $p(v_t \mid v_{t-1}, \alpha, f_t)$, where $\alpha$ and $f_t$ are the Dirichlet prior and the frequency counts used for the posterior over the map structure at time $t$. These modifications yield the following, more simple update rule for locations outside the partial map.

---

[3]Obviously, the transitions between $v_{t-1}$ and $v_t$ also depend on how far the robot moved. In our current implementation we update the view whenever the robot moved two meters, which makes the transition probabilities sufficiently stable.

$$p(x_t \mid z_{1:t}, u_{1:t-1}) \propto \sum_{v_t, v_{t-1}} \int p(z_t \mid v_t)\, p(v_t \mid v_{t-1}, \alpha, f_t) \cdot$$
$$p(x_t \mid x_{t-1}, u_{t-1})\, \bar{p}(x_{t-1})\, \bar{p}(v_{t-1})\, dx_{t-1} \quad (7)$$

To summarize, the key idea of our approach to map merging is to sequentially estimate a robot's location both inside and outside the partial map built by the other robot. Locations inside the map are updated using (5) and locations outside the map are updated based on (7). To estimate the likelihood of observations outside the map, the technique estimates a structural parameter $q_t$ along with the robot's location. At each iteration, this parameter is updated using the frequency counts based on the most likely views extracted from the observations.

## 4 Implementation

### 4.1 Particle filter for partial map localization

The generative model for map merging is implemented using a particle filter [6, 3]. A detailed description of this implementation can be found in [9]. Particle filters represent posteriors over a robot's continuous position by sets $S_t = \{\langle x_t^{(i)}, w_t^{(i)} \rangle \mid i = 1, \ldots, N\}$ of $N$ weighted samples distributed according to the posterior. Here each $x_t^{(i)}$ is a sample (or state), and the $w_t^{(i)}$ are non-negative numerical factors called *importance weights*, which sum up to one. Sets at time $t$ are generated from previous sets $S_{t-1}$ by a sampling procedure often referred to as SISR, sequential importance sampling with re-sampling [3]. SISR implements the recursive Bayes filter update rule (5) in a three stage process: First, draw states $x_{t-1}^{(i)}$ from the previous sample set with probability given by the importance weights $w_{t-1}^{(i)}$, then draw for each such state a new state from the predictive distribution $p(x_t \mid x_{t-1}^{(i)}, u_{t-1})$, and finally weight these new states/samples proportional to the observation likelihood $p(z_t \mid v_{x_t})$. The last step, importance sampling, adjusts for the fact that samples are not drawn from the actual posterior distribution but from the predictive distribution.

The generative model for map merging described in the previous section requires to estimate the posterior over robot locations both inside and outside the partial map. We assume that the size of the area outside the partial map can be set based on an estimate of the total size of the environment. Clearly, a representation of all locations outside the map would require too many samples for online estimation. Our solution to this problem is based on the idea that, along with its history, a sample can be seen as the end point of a robot trajectory. This allows us represent only those samples (trajectories) for which the robot was inside the partial map at some point in time. To do so, our approach initially generates samples uniformly distributed inside the partial map. At later iterations, samples enter and exit the map, depending on their location and the robot's motion (see [9]). At each iteration, the samples inside the map are weighted by $p(z_t \mid v_{x_t})$, i.e. likelihood of the observation given the robot's position in the partial map. All samples outside the partial map are weighted by $p(z_t \mid \text{outside})$, the likelihood of the observation computed from the structural model:

$$p(z_t \mid \text{outside}) \propto \sum_{v_t, v_{t-1}} p(z_t \mid v_t)\, p(v_t \mid v_{t-1}, \alpha, f_t) \bar{p}(v_{t-1}) \quad (8)$$

This term for the importance weight follows directly from (7). The structural parameter $q_t$ and the distribution over views $v_t$ is updated as described in Section 2. After each iteration, the samples represent a robot's location relative to the partial map built by another robot. Each sample along with its history represents a unique match between the partial maps built by the two robots. Once a match with sufficiently high probability is found, map merging can be performed by a mapping algorithm such as [14]. The partial map localization algorithm is highly efficient and can be computed in real time on a state-of-the-art laptop.

### 4.2 View extraction

To test our approach using data collected by real robots we have to extract discrete views from sensor data. Since this is not the current focus of our work, we implemented a rather simple technique that extracts structural information from laser range-scans. To do so, the approach sequentially evaluates the individual beams of a laser scan and checks for differences between neighboring beams. Depending on their relationship, consecutive beams are clustered into groups denoted w, g, m, and c. Group w (for wall or flat obstacle) is assigned to groups of beams for which all neighboring beams measure similar distances, g for large gaps between two beams, m for max range readings, and c for corners (based on lines extracted from the scan). Thus each laser scan is represented as a string of these four letters. Fig. 4 shows two example laser scans along with the corresponding feature strings (counterclockwise).

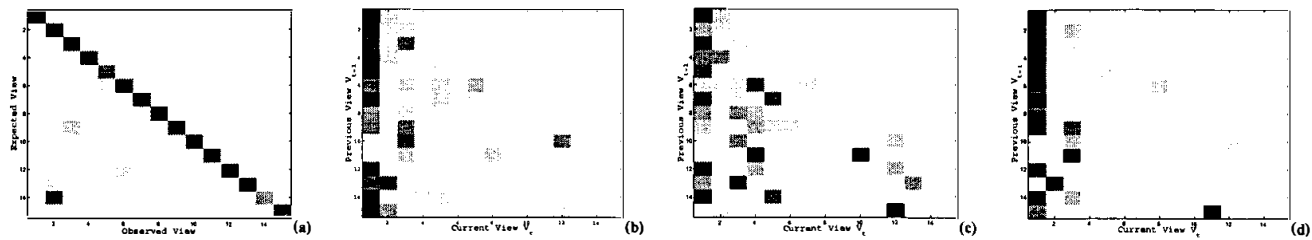

Figure 5: Learned models: (a) Observation model. (b)–(d) View transitions $q$. The $y$-axes represent views $v_{t-1}$ and the $x$-axes give the following view $v_t$. Shown are only the 15 most frequent views, higher probabilities are darker. (b) Prior $\alpha$ extracted from all maps, (c) posterior for map 4, and (d) posterior for map $k$ in Fig. 2.

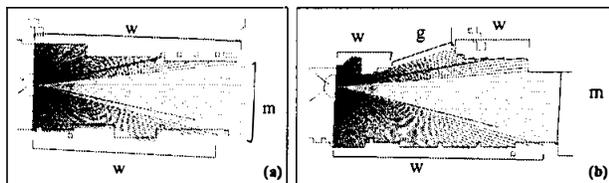

Figure 4: Two laser scans corresponding to the most frequently observed views. The robot is on the left side moving to the right. (a) wmw, is typically observed when a robot moves down a hallway. (b) wmwgw, indicates that the robot approaches an opening (gap) on its left.

The key advantage of this model is that it is extremely robust in capturing the main structural elements of an environment such as hallways, junctions, rooms, and corners. Furthermore, the detected features are robust with respect to rescaling (*e.g.* different widths of hallways). A disadvantage of these views is that they do not provide accurate location information. We overcome this problem by weighting samples inside the partial map using the raw laser scans, which provide highly accurate location information [6]. This technique has no impact on our solution to the revisiting problem, since these samples are still weighted against samples outside using the views as described in the previous section.

The parameters of the model were hand-tuned so as to get satisfying results. After merging symmetric views, in the 35,000 laser scans collected in the environments shown in Fig. 3, only $\nu = 37$ different scan "strings" occurred.

## 5 Experiments

The experiments were carried out using data collected in the five environments shown in Fig. 3.

### 5.1 Learning structural models

To learn structural models, we used 35,000 pairs of consecutive views (strings) collected by mobile robots when mapping the different environments. The parameters of the learned modes are shown in Fig. 5. Each graph plots the probability matrix for the 15 most frequent views. These 15 views cover approximately 80% of all observed scans. View 1 and 2 are the strings wmw and wmwgw, illustrated in Fig. 4(a) and (b), respectively. Fig. 5(a) shows the observation model $p(z_i|v_j)$ extracted from the data. This model was learned using the same hierarchical approach as the one described in Section 2.2 for map structures. In this context the hyperparameters smooth the extracted counts of $p(z_i|v_j)$. The high probabilities on the diagonal indicate that our view extraction is very robust. The prior transition model $\alpha$ extracted from all maps is shown in Fig. 5(b). Not surprisingly, most views have a high probability to transition to the hallway view 1, since the training environments contain many long corridors. When comparing the posteriors shown in Fig. 4(c) and (d), it becomes clear that the approach was able to extract the fact that environment 4 has far less hallways than environment $k$ in Fig. 2.

### 5.2 Partial map localization

We systematically evaluated our approach to map merging under global uncertainty using the following scenario. Imagine two robots are placed at random locations in an unknown environment. Both robots start to explore the environment and at some point they can communicate. At that point, one robot localizes the other robot in its own map so as to determine whether there is an overlap between the two maps. We generated 15 partial maps based on data collected in the three environments labeled 2, 4, and $k$ in Fig. 2. Some of these maps are shown in Fig. 6. In our scenario, one robot used these partial maps to localize the other robot based on data collected in the same environment. For each environment, we generated a prior structural model $\alpha$ based on the other environments only. The data of the other robot consisted of 25 data sequences for each environment, resulting in a total of $5 \cdot 25 = 125$ map-trajectory pairs for each environment. The results given below are averaged over the average performances in the three different environments. For each pair we proceeded as follows. Let A denote the robot with the partial map and B the other robot. At each iteration of the particle filter, robot A determines the probability of the most likely hypothesis for B's position in its map. A considers a hypothesis to be valid if its probability exceeds a certain threshold $\theta$.

The solid line in Fig. 7 shows the resulting precision-recall trade-off for different thresholds using our approach (each point on the line represents a different threshold). For each threshold $\theta$, *precision* measures the fraction of the correct matches among those hypotheses that are considered valid, i.e. above the threshold. Correctness is tested by comparing the position of the hypothesis to a ground truth estimate computed offline. To determine recall, we first checked at what times robot B was in robot A's partial map. *Recall*, then, measures the fraction of this time for which robot A generated a correct hypothesis, i.e. at the correct position and with probability above the threshold $\theta$. To show the advantage of our approach, we compared it to an alternative method that uses a *fixed* likelihood $p(z_t \mid \text{outside})$ for locations outside the partial map (compare to (8) for our approach). The trade-offs resulting from different fixed likelihoods are plotted as dashed lines in Fig. 7 (data points are omitted for clarity). The graph clearly shows the superior performance of our approach. It achieves 26% higher precision than the best likelihood value for the alternative method. Note that high precision values are more important than high recalls since low precision results in wrong map merges while low recall only delays the map merging decision. Note also that one cannot expect very high recall values since robot B has to be in the partial map for a certain duration before a valid hypothesis can be generated.

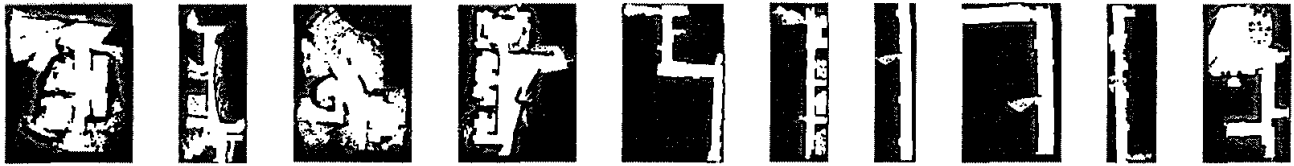

Figure 6: Partial maps used for evaluation of map merging. The maps were taken from three different environments. In each experiment, one robot built such a partial map and receives data from another robot collected in the same environment. The robot has to determine when and if so where the other robot is in its partial map.

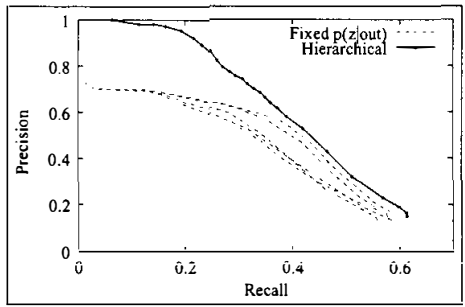

Figure 7: Precision vs. recall: Each point represents an average over 375 pairs of partial maps and trajectories. Each curve shows the trade-off for different thresholds $\theta$ (0.05-0.99). The dashed lines indicate results obtained with different fixed values for $p(z \mid \text{outside})$ and the solid line represents the results of our approach.

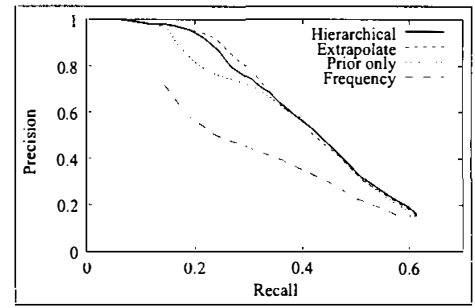

Figure 8: Precision vs. recall for different adaptive techniques. The black, dashed line is obtained when using pure frequency counts for the individual features.

Fig. 8 shows the same evaluation for different ways of updating and learning map structures. The dashed line denoted by "Frequency" represents the results obtained without considering the transition model for views. This approach uses frequency counts obtained from the training maps to compute the likelihood $p(z_t \mid \text{outside})$ of a view. This likelihood is computed by dividing the number of times this view was observed in the training data by the total number of observations. The bad performance of this method confirms our belief that it is crucial to consider the *connective structure* of environments as modeled by our Dirichlet process. The dotted line represents the results obtained without updating the structural parameter $q$ during map merging, i.e. $q_t$ is set to the prior $\alpha$. It can be seen that adjusting the estimation process during map merging increases the robustness of the approach. Finally, the short dashed, best curve shows a variant of our hierarchical approach that weights the observed frequencies proportional to the ratio between the size of the partial map and the size of the entire environment. In essence, this approach extrapolates the observations made in the partial map assuming that the unexplored areas have the same structure.

In these experiments we only tested the quality of the estimation process underlying the decision problem in multi-robot map merging. Our current project aims to field 100 robots in an indoor exploration and reconnaissance task. To achieve maximum robustness against false positive map merges, our multi-robot control system additionally verifies the hypotheses generated by the partial localization approach described here. Robots verify a match hypothesis by meeting at a location that follows from the hypothesis. The integration of this approach into a decision-theoretic robot exploration strategy is described in [9].

## 6 Conclusions and Future Work

In this paper, we introduced a novel approach to addressing the revisiting problem in mobile robot map building. Multi-robot map merging, a particularly difficult instance of this problem, requires the localization of one robot relative to a partial map built by another robot. The key problem in map merging without knowledge about the robot's relative locations is to get accurate estimates for the likelihoods of observations *outside* the partial map. To solve this problem, we introduce a structural model of an environment that can be used to predict the observations made by the robot. The structural model is a hidden Markov model that generates sequences of views observed by a robot when navigating through the environment. The parameters of the model are updated during exploration via Dirichlet hyperparameters. A Dirichlet prior is learned from previously encountered environments.

The structural model is integrated into a particle filter that uses samples to represent a robot's location and that updates the structural parameters as more data becomes available. Extensive experiments show that our approach significantly outperforms an alternative technique that uses a fixed likelihood for observations outside the partial map. We were not able to find a likelihood that yielded results comparable to our method.

The approach presented here can be readily applied to the loop closing problem in single robot mapping (see Fig. 1(b)). Here, a robot has to decide whether it came back to a previously explored location, or whether it moves through a similar, unexplored area. Especially mapping approaches based on Rao-Blackwellised particle filters [2, 12, 4] can easily incorporate our structural model. Just like in multi-robot map merging, the model can then be used to assign appropriate probabilities to location hypotheses (particles) in unexplored areas.

Despite these encouraging results, this is only the first step towards using structural models of environments. For example, our current approach uses maximum likelihood estimates to update the parameters of the model. More sophisticated EM-based techniques such as [1, 16] might yield further improvements. Other areas for improvement are better algorithms for extracting views from sensor data. Another application of our method is to improve robot exploration strategies by predicting partial maps into unexplored areas. Thereby, for example, a robot can actively try to close loops so as to improve map quality.

We consider hierarchical Bayesian techniques such as the one used in this paper to be an extremely promising tool for achieving more robust estimation and reasoning processes in robotics. Most existing approaches to state estimation in robotics are fixed in that they do not adapt to the environment. For example, if a map building approach is based on the assumption that the environment is rectilinear it will fail in environments that violate this assumption. On the other hand, not making use of the fact that most environments are rectilinear obviously discards valuable information. Using a hyperparameter modeling the type of environment, a mapping approach would work reliably in different types of environments while still being able to make use of the structure underlying a specific environment.

**Acknowledgments**

This work has partly been supported by the NSF under grant number IIS-0093406 and by DARPA's SDR and MICA programs (contract numbers NBCHC020073 and AFRL F30602-01-C-0219). The authors would also like to thank Aaron Hertzmann, Geoff Gordon, and Sebastian Thrun for stimulating discussions.